\documentclass[letterpaper]{article}

\usepackage{aaai18}
\usepackage{times}
\usepackage{helvet}
\usepackage{courier}
\usepackage{color}
\usepackage{array,booktabs,multirow}
\usepackage{amsmath,amssymb,amsthm}
\usepackage{latexsym}
\usepackage{amsfonts}
\usepackage{xspace}
\usepackage{graphics}
\usepackage{multirow}
\usepackage{graphicx}
\usepackage{todonotes}
\usepackage{url}
\usepackage[all]{xy}

\newtheorem{example}{Example}

\newtheorem{proposition}{Proposition}

\theoremstyle{definition}
\newtheorem{definition}{Definition}

\newcommand{\bag}{\ensuremath{\mathcal{\textbf{A}}}}

\newcommand{\weight}{\ensuremath{\mathcal{\operatorname{w}}}}
\newcommand{\arguments}{\ensuremath{\mathcal{A}}}
\newcommand{\attacks}{\ensuremath{\mathcal{R}}}
\newcommand{\supports}{\ensuremath{\mathcal{S}}}
\newcommand{\diff}[1]{\ensuremath{\frac{\mathrm{d}#1}{\mathrm{d}t}}}

\newcommand{\attacker}{\ensuremath{\mathrm{Att}}}
\newcommand{\supporter}{\ensuremath{\mathrm{Sup}}}
\newcommand{\energysolution}{\ensuremath{\sigma^\bag}}

\newcommand{\iterscheme}{\ensuremath{\mathcal{I}}}

\title{A Tutorial for Weighted Bipolar Argumentation with Continuous Dynamical Systems and the Java Library Attractor}
\author{Nico Potyka\\
           Institute of Cognitive Science, University of Osnabr\"{u}ck, Germany}

\nocopyright

\begin{document}

\maketitle
% \addtolength{\abovedisplayskip}{-1mm}
% \belowdisplayskip1.8mm

\begin{abstract}
Weighted bipolar argumentation frameworks allow modeling decision problems and
online discussions by defining arguments and their relationships.
The strength of arguments can be computed based on an initial weight and
the strength of attacking and supporting arguments.
While previous approaches assumed an acyclic argumentation graph and successively 
set arguments' strength based on the strength of their parents,
recently continuous dynamical systems have been proposed as an alternative.
Continuous models update arguments' strength simultaneously and continuously. 
While there are currently no analytical guarantees for convergence
in general graphs, experiments show that continuous models can converge 
quickly in large cyclic graphs with thousands of arguments. 
Here, we focus on the high-level ideas of this approach
and explain key results and applications.
We also introduce \emph{Attractor}, a Java library that can be used to solve 
weighted bipolar argumentation problems.
\emph{Attractor} contains implementations of several discrete and continuous models and numerical algorithms to compute solutions.
It also provides base classes that can be used 
to implement, to evaluate and to compare continuous models easily.
\end{abstract}

\section{Introduction}

\emph{Abstract argumentation} \cite{dung1995acceptability} studies the acceptability
of arguments based purely on their relationships and abstracted from their content. 
The basic framework has a two-valued semantics and allows only defining arguments and an attack relation between them.
This basic setting has been extended in different directions.
For example, \emph{bipolar argumentation frameworks} \cite{amgoud2004bipolarity,OrenN08,cayrol2013bipolarity,polberg2014revisiting} take account of the fact that arguments cannot 
only attack each other and add a support relation. A survey of different approaches can be found in \cite{DBLP:journals/ker/CohenGGS14}. The classical two-valued semantics
that distinguishes only between acceptance and rejection of arguments
has been extended in various ways. Examples include probabilistic semantics \cite{thimm2012probabilistic,hunter2013probabilistic,HunterPotyka17} and 
ranking semantics that can be based on fixed point equations \cite{besnard2001logic,leite2011social,correia2014efficient,barringer2012temporal} or 
the graph structure \cite{cayrol2005graduality,amgoud2013ranking}.
Other recent extensions include recursive attacks on attacks \cite{baroni2011afra}
and the temporal availability of arguments \cite{budan2015modeling}.

Our focus here is on \emph{weighted bipolar argumentation frameworks} that allow defining attack and support relationships and an initial weight for arguments \cite{baroni2015automatic,rago2016discontinuity,amgoud2017evaluation,mossakowski2018modular}.
A strength value is computed for every argument based on its initial weight and the 
strength of its attackers and supporters. 
Examples for computational models include 
the \emph{QuAD algorithm} from \cite{baroni2015automatic} that was designed to evaluate the strength
of answers in decision-support systems. Soon after, the \emph{DF-QuAD algorithm} \cite{rago2016discontinuity} was proposed as an alternative, which avoids discontinuous behaviour of 
the QuAD algorithm that can be undesirable in some applications. Some additional interesting guarantees are given by the \emph{Euler-based semantics} introduced in \cite{amgoud2017evaluation}.
The QuAD algorithms mainly lack these properties due to the fact that their aggregated strength values \emph{saturate}.
That is, as soon, as an attacker (supporter) with strength $1$ exists, the other attackers (supporters) become irrelevant for the aggregated value.
However, while the Euler-based semantics avoids these problems, it has some other drawbacks that may be undesirable. Arguments initialized with strength $0$ or $1$ remain necessarily unchanged
under Euler-based semantics and the impact of attacks and supports is non-symmetrical. The \emph{quadratic energy model} introduced in \cite{potyka2018Kr} avoids these problems.
In particular, while the previous approaches are discrete in nature, the quadratic energy model is a continuous model. 
Discrete models often assume that the argumentation graph is acyclic, so that the strength of arguments 
can be computed successively according to a topological ordering.
Continuous models change arguments' strength continuously and simultaneously. They can be naturally applied to cyclic graphs, but convergence in general remains an open question.

More formally, continuous models correspond to n-dimensional functions $f(t)$ mapping continuous points in time
to $n$-dimensional state vectors whose $i$-th component represents the strength of the $i$-th argument 
at time $t$. The initial state $f(0)$ is given by the initial weights and a system of differential equations
describes how the strength values evolve as time progresses.
This approach, in particular, allows plotting the evolution of strength values in order to better
understand the final strength values of the limit $s = \lim_{t \rightarrow \infty} f(t)$
or to inspect the convergence behaviour visually.
Even though convergence in general graphs is an open question, 
so far no diverging example has been found
and tests with large randomly generated bipolar graphs show that the strength values converge in many cases.

In this tutorial paper, we will review some ideas and results from \cite{potyka2018Kr}
with a stronger focus on the high-level ideas. The goal of this paper is, in particular,
to demonstrate how results can be applied to
\begin{enumerate}
  \item solve weighted bipolar argumentation problems with continuous models,
	\item transform existing discrete models to well-defined continuous models.
\end{enumerate}
We also introduce \emph{Attractor}, a Java library that provides basic implementations of the ideas
discussed here and in \cite{potyka2018Kr}. 
The main goals of \emph{Attractor} are to
\begin{enumerate}
	\item simplify applying continuous models to weighted bipolar argumentation problems,
	\item to improve reproducibility of the results in \cite{potyka2018Kr},
	\item to simplify implementing new continuous models and
	\item to simplify comparing different models.
\end{enumerate}
These goals are achieved by providing 
\begin{enumerate}
	\item implementations of some continuous models and the random generator introduced in \cite{potyka2018Kr},
	\item implementations of base classes that solve initial value problems with basic and advanced methods,
	\item implementations of utility classes for benchmarking, plotting and working with benchmark files. 
\end{enumerate}
We will start with an introduction to dynamical systems and the quadratic energy model from \cite{potyka2018Kr} in Section
\ref{sec_dyn_systems}. In Section \ref{sec_computation}, we will discuss the problem of computing solutions and 
explain some important algorithms. Section \ref{sec_convergence} contains some additional information
on convergence guarantees and open questions and discusses the computational complexity of the continuous approach.
In Section \ref{sec_discrete_continuous}, we explain how discrete models can be transformed to continuous models.
Finally, Section \ref{sec_attractor} explains how the previously discussed ideas can be put into practice using \emph{Attractor}.

\section{Dynamical Systems and The Quadratic Energy Model}
\label{sec_dyn_systems}

Roughly speaking, a dynamical system describes the evolution of a natural or technical system over time. 
If time is discretized, the system is called discrete, otherwise it is called continuous.
Formally, we describe the state of the system at time $t$ by a function $s(t)$.
The state is usually given as a real vector and a system of differential equations describes how the
system evolves dependent on the current state.

In the context of weighted argumentation, a state vector contains one component for every argument that can take a strength value between $0$ and $1$.
The strength of arguments should evolve based on the initial weight, and the current strength of attackers and supporters.
Before describing this approach in more detail, we define \emph{weighted bipolar argumentation graphs (BAGs)} as introduced in \cite{amgoud2017evaluation}.
\begin{definition}[BAG]
A \emph{BAG} is a quadruple $\bag = (\arguments, \weight, \attacks, \supports)$, where
$\arguments$ is a finite set of arguments, $\weight: \arguments \rightarrow [0,1]$ is a weight function and $\attacks$ and $\supports$ are binary relations on $\arguments$ called \emph{attack} and
\emph{support}.
\end{definition}
In order to simplify notation, we  assume  that the $i$-th argument is  called $i$, 
that is, $\arguments = \{1, \dots, n\}$.
The weight function $\weight$ defines an initial strength value between $0$ and $1$ for each argument. If $a \attacks b$ ($a \supports b$), we say that $a$ attacks (supports) $b$.
We let $\attacker_i = \{h \in \arguments \mid h \attacks i\}$ denote $i$'s attackers and
let $\supporter_i  = \{h \in \arguments \mid h \supports i\}$ denote $i$'s  supporters. 

\begin{figure}[tb]
	\centering
		\includegraphics[width=0.46\textwidth]{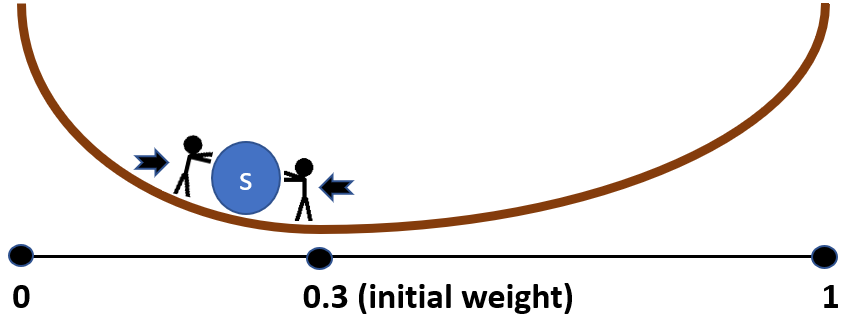}
			\caption{Illustration of dynamical system.
			\label{fig_physical_metaphor}}
\end{figure}

We can now describe our dynamical system more precisely.
A state is a vector $s \in \mathbb{R}^n$ whose $i$-th component $s_i$ is the strength of argument $i$.
Our state model is a function $s: \mathbb{R}^+_0 \rightarrow \mathbb{R}^n$ that maps non-negative 
time points to strength vectors. That is, $s_i(t)$ is the strength of argument $i$ at time $t$.
Initially, the strength of an argument should correspond to its initial weight, that is, we let $s_i(0) = \weight(i)$.
The evolution of the system should be based on three considerations:
\begin{enumerate}
	\item Strength values are attracted by their initial weight.
	\item Attackers force the strength value towards 0 proportionally to their strength.
	\item Supporters force the strength value towards 1 proportionally to their strength.
\end{enumerate}
Intuitively, there are three forces acting on the strength of each argument as illustrated in
Figure \ref{fig_physical_metaphor}. In this physical metaphor,
attackers push the strength towards $0$, while 
the supporters push the strength towards $1$. 
Gravity pulls the strength back to its initial weight.

This intuition can be modeled by a system of differential equations. If this system is designed carefully,
it uniquely defines a model $s_\bag: \mathbb{R}^+_0 \rightarrow \mathbb{R}^n$ for every BAG $\bag$.
We are then interested in the long-term behaviour of the model.
Intuitively, we expect the forces to counteract until the strength reaches an equilibrium state
where all forces are in balance. More formally,
if the model converges to a state $s^* = \lim_{t \rightarrow \infty} s(t)$ as time progresses, 
we call $s^*$ the \emph{equilibrium state} reached by the model. 

The quadratic energy model introduced in \cite{potyka2018Kr} is defined as follows.  
\begin{definition}[The Quadratic Energy Model]
\label{def_quadratic_energy_model}
Let $\bag$ be a BAG.
For all $j \in \arguments$, the \emph{energy at $j$} is defined as 
$$E_j = \sum_{i \in \supporter_j} s_i - \sum_{i \in \attacker_j} s_i$$
and for all $x \in \mathbb{R}$, the \emph{impact of $x$} is defined as
$$h(x) = \frac{\max\{x,0\}^2}{1 + \max\{x,0\}^2}.$$ 
The \emph{quadratic energy model $\energysolution: \mathbb{R}^+_0 \rightarrow \mathbb{R}^n$} for $\bag$ is the unique solution of the system of differential equations
\begin{align}
\label{eq_energy_flow_des}
\diff{s_j} = \weight(j) - s_j 
&+  (1 - \weight(j)) \cdot h(E_j) \notag\\
&-  \weight(j) \cdot h(-E_j), \qquad j \in \arguments 
\end{align}
with initial conditions 
$s_j(0) = \weight(j)$.
\end{definition}
Intuitively, $\diff{s_j}(t)$ describes the momentary change of $s_j$ at time $t$. 
If $\diff{s_j}(t) > 0$, the strength will increase, if $\diff{s_j}(t) < 0$, the strength will decrease and if $\diff{s_j}(t) = 0$ the strength will remain in its current state.
The definition uses two auxiliary functions. The \emph{energy} $E_j$ at argument $j$ aggregates the strength of attackers and supporters in a linear fashion.
This notion of energy has been first defined for the Euler-based restricted semantics 
in \cite{amgoud2017evaluation}.
The energy is then fed into the \emph{impact} function $h$ that is $0$ for all negative arguments and then strictly increases, but is bounded from above by $1$.
The definition of $\diff{s_j}$ can be divided into three parts that correspond to our
three considerations above.
\begin{enumerate}
	\item The difference $\weight(j) - s_j$ draws the strength of an argument to its initial weight.
	Notice that if $\weight(j) > s_j$ ($\weight(j) < s_j$), this term is positive (negative) and tends to increase (decrease) $j$'s strength.
	\item The term $- \weight(j) \cdot h(-E_j)$ moves the strength towards $0$ if the negative force of attackers is stronger than the positive force
	of supporters.
	\item Dually, the term $(1 - \weight(j)) \cdot h(E_j)$ moves the strength towards $1$ if the positive force of supporters is stronger than the negative force
	of attackers.
\end{enumerate}
As shown in \cite{potyka2018Kr}, the quadratic energy model is well-defined. That is,
the system has a unique solution $\energysolution$ by means of which we can simulate the evolution of strength values over time. 
If $\energysolution$ reaches an equilibrium state, the final strength values at every argument
are completely determined by the energy at this state as explained in the following proposition.
\begin{proposition}[Strength in Equilibrium \cite{potyka2018Kr}]
\label{prop_energysolution_equilibrium}
If the limit $s^* = \lim_{t \rightarrow \infty} \energysolution(t)$ exists, then we have 
\begin{align}
\label{eq_energysolution_limit_equation}
s^*_j = 
\begin{cases}
\weight(j) & \text{if } E_j = 0\\
\weight(j) +  (1 - \weight(j)) \cdot h(E_j)   & \text{if } E_j > 0\\
\weight(j) - \weight(j) \cdot h(-E_j).  & \text{if } E_j < 0
\end{cases}
\end{align}
\end{proposition}
Equation \ref{eq_energysolution_limit_equation} shows, in particular, that the strength will be the
initial weight if the energy is $0$ and otherwise will go to $1$ ($0$) as the energy goes to $\infty$ ($-\infty$).
The quadratic energy model satifies a collection of postulates proposed in \cite{amgoud2017evaluation}.
These postulates range from very general properties like \emph{Anonymity} (strength values do not depend on the identity of the argument) and \emph{Independence} (arguments are independent 
of disconnected arguments) to properties tailor-made for weighted bipolar argumentation frameworks that guarantee that attacks, supports and initial weights
have the intended meaning. An interesting property that distinguishes weighted argumentation frameworks from some other numerical argumentation frameworks is \emph{Directionality},
which guarantees that arguments influence other arguments only in direction of the edges. This property distinguishes weighted argumentation approaches from probabilistic approaches
that usually cause influence in both directions due to the nature of probability theory.
Please see \cite{amgoud2017evaluation} and \cite{potyka2018Kr} for a more thorough discussion of the properties.

In the following example, we illustrate the quadratic energy model by means of a small
decision problem.
\begin{example}
\label{exp_phone_stocks}
Suppose we want to decide whether to buy or to sell stocks of an electronics company.
We base our decision on information given by different experts:
\begin{description}
	\item[1:] The development of the new phone was too expensive. Therefore, the company has to cut down research and development and will not stay competitive in the future.
	\item[2:] The company's new phone is innovative and will increase the company's profit considerably.
	\item[3:] The price of the new phone is too  high and there will not be too many sales.
	\item[4:] There is a large number of preorders of the new phone already.
	\item[5:] The company's investment in research and development is far beyond competitors' investment and the company is likely to become the market leader in the future.
\end{description}
Initially, we do not have a preference for buying or selling stocks and 
set both initial weights to $0.5$. In order to weigh the expert opinions,
we could use historical information about how frequently the expert's assessment
was true or false. If there were $t$ true and $f$ false assessments,
we could set the initial weight to $\frac{t}{t+f}$. In order to incorporate
arguments of new experts and to set an initial bias for the weight, 
we could add pseudocounts to $t$ and $f$.
That is, we set the initial weight to $\frac{t+t'}{t+t'+f+f'}$, where $t', f' \in \mathbb{N}$
are pseudocounts that encode an initial bias. If $t'=f'$, our initial weight is $0.5$ when no historical information is available.
Setting $t'>f'$ ($t'<f'$), the weight will initially be greater (smaller) than $0.5$.
The larger $t'+f'$ is, the more data is needed to deviate from the bias.
For instance, if $t=5$, $f=2$, then we have the weight $\frac{5+1}{5+1+2+1} \approx 0.66$
for pseudocounts $t'=f'=1$, whereas we have $\frac{5+10}{5+10+2+10} \approx 0.55$
for pseudocounts $t'=f'=10$. Figure \ref{fig_stock_example_bag}
shows a BAG for our problem along with initial weights and the final strength
values under the quadratic energy model.
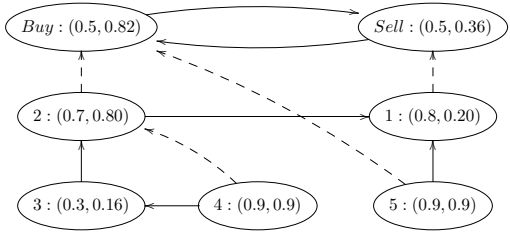
\begin{figure}[tb]
		  \scalebox{.65}{
				\xymatrix{
			%   Line1
  *+++[o][F-]{Buy: (0.5,0.82)} \ar@{->}@/^1pc/[rr] && *+++[o][F-]{Sell: (0.5,0.36)} \ar@{->}@/^1pc/[ll] \\
	     *+++[o][F-]{2: (0.7,0.80)} \ar@{-->}[u] \ar@{->}[rr] && *+++[o][F-]{1: (0.8,0.20)} \ar@{-->}[u] \\
			*+++[o][F-]{3: (0.3,0.16)} \ar@{->}[u] & *+++[o][F-]{4: (0.9,0.9)} \ar@{-->}@/_1pc/[ul] \ar@{->}[l] &  *+++[o][F-]{5: (0.9,0.9)} \ar@{->}[u] \ar@{-->}@/_1pc/[uull]
			}	
		}
		\caption{BAG for stock examples. Nodes show (initial weight, final strength) under
		quadratic energy model. \label{fig_stock_example_bag}}
\end{figure}
One advantage of a continuous model is that we can illustrate
the evolution of the strength values by drawing function graphs for selected arguments over time. This makes it easier to explain the final strength values.
\begin{figure}[tb]
	\centering
		\includegraphics[width=0.46\textwidth]{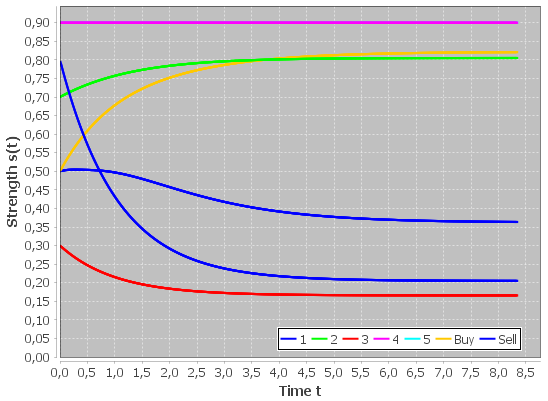}
			\caption{Long-term behaviour of $\energysolution$ for BAG in Figure \ref{fig_stock_example_bag}.
			\label{fig_stock_example_graph}}
\end{figure}
Recall that a state in our dynamical system contains a strength value for every argument. 
Our example is sufficiently small to draw all function graphs simultaneously
without making the picture too messy. The graphs are shown in Figure \ref{fig_stock_example_graph}. For example, the blue graph starting from $0.5$ shows the evolution of the
strength of the sell-argument. Initially, its strength slightly increases due to the support
by argument $1$ (the blue graph starting from $0.8$). However, argument $1$ becomes gradually
weaker due to its attackers. At about time $0.5$, the attacking buy-argument becomes as strong as argument $1$ (the blue and yellow graphs intersect) and the strength of the selling-argument
starts decreasing. It actually starts decreasing slightly before that because it is also drawn to its initial weight $0.5$.  
\end{example}

\section{Numerical Computation of Solutions}
\label{sec_computation}

Even though we know that the quadratic energy model  $\energysolution$ is well-defined
for every BAG, we usually cannot solve for $\energysolution$ analytically.
This is the case for most nonlinear dynamical systems, but is not a heavy drawback in practice
since the solution can be approximated numerically.
 
The intuitive idea is best illustrated by \emph{Euler's method}, which is a simple
algorithm to perform this approximation.
Recall that $\diff{s_j}(t)$ describes the momentary change of $s_j$ at time $t$.
Hence, if we know $s_j(t)$, we can approximate $s_j(t + \delta)$ by letting
$\hat{s}_j(t + \delta) = s_j(t) + \delta \cdot \diff{s_j}(t)$.
Formally, this approach is justified by the fact that differentiable functions
can be approximated locally by the derivative. In particular, as we let the step size
$\delta$ go to $0$, the approximation error $|\hat{s}_j(t + \delta) - s_j(t + \delta)|$ goes to $0$ as well.

In the context of weighted bipolar argumentation, we can initialize all
strength values with the initial weights. This gives us $s(0)$ and we can compute
$\diff{s_j}(0)$ according to Definition \ref{def_quadratic_energy_model}. We can then approximate $\energysolution(\delta)$ by letting
$\hat{s}(\delta) = s(0) + \delta \cdot \diff{s}(0)$. Given our approximation 
$\hat{s}(\delta)$, we can compute $\diff{\hat{s}}(\delta)$ and 
$\hat{s}(2 \cdot \delta) \approx \hat{s}(\delta) + \delta \cdot \diff{\hat{s}}(\delta)$.
Continuing in this way, we can compute $\hat{s}(n \cdot \delta)$ for arbitrary 
$n \in \mathbb{N}$. Hopefully, the strength values will eventually converge. 
This is the case if and only if the derivative $\diff{\hat{s}}(t)$ goes to $0$.
Therefore, a simple termination condition is to demand that $\diff{\hat{s}_i}(t) \leq \epsilon$
for all $i \in \arguments$ and some small $\epsilon > 0$.
Formally, this corresponds to demanding that the maximum norm of $\diff{\hat{s}}(t)$
is smaller than $\epsilon$, denoted as $\| \diff{\hat{s}} \|_\infty \leq \epsilon$.
The complete algorithm is shown in Figure \ref{fig:EulersMethodAlgorithm}.
\begin{figure}
	\begin{tabular}{l}
		\textit{EulerApproximation}($\bag$, $\delta$, $\epsilon$): \\[0.1cm]
		\hspace{0.2cm} $t \leftarrow 0$ \\[0.1cm]
		\hspace{0.2cm} \textit{for} $i \in \arguments$: \\[0.1cm]
		\hspace{0.4cm} $s_i \leftarrow \weight(i)$ \\[0.2cm]
		\hspace{0.2cm} \textit{while} $\| \diff{s} \|_\infty > \epsilon$: \\[0.1cm]
		\hspace{0.4cm} \textit{for} $i \in \arguments$: \\[0.1cm]
		\hspace{0.6cm} $\quad s'_i \leftarrow s_i + \delta \cdot \diff{s_i}$ \\[0.2cm]
		\hspace{0.4cm} $s \leftarrow s'$ \\[0.2cm]
		\hspace{0.2cm} \textit{return} $s$
	\end{tabular}
	\caption{Euler's method for approximating the energy model $\energysolution$ given a 
	BAG $\bag = (\arguments, \weight, \attacks, \supports)$, step size $\delta$ and 
	convergence threshold $\epsilon$.}
	\label{fig:EulersMethodAlgorithm}
\end{figure}

While Euler's method is easy to understand and to implement, it does not give very strong
approximation guarantees.
A better alternative that is frequently used is the family of Runge–-Kutta methods. 
The most prominent member is the classical Runge–-Kutta method RK4 that guarantees
an approximation error in the order of $O(\delta^4)$. In practice, this means that if we halve the step size $\delta$ 
(double the number of iterations), we usually decrease the approximation error by a factor 
of $16$ \cite{polyanin2017handbook}.
Since the derivatives of the quadratic energy model 
can never become larger than $1$, using RK4 with constant step size $0.01$
should be safe.
If we want to make sure that the step size is sufficiently small, we can run the algorithm
until termination and then repeat with a smaller step size like $0.005$ and check
that the final values remain unchanged up to the desired accuracy.

\section{Convergence and Complexity}
\label{sec_convergence}

The quadratic energy model $\energysolution(t)$ uniquely defines the evolution of
strength values for every BAG $\bag$ over time.
Our hope is that $\energysolution(t)$ converges to an equilibrium state as $t \rightarrow \infty$.
This allows us to define strength values by means of the equilibrium $s^* = \lim_{t \rightarrow \infty} \energysolution(t)$.
Unfortunately, convergence of $\energysolution(t)$ for general BAGs with cycles 
is currently an open question. One can show convergence for some special cases.
For example, if cycles contain only support relations, the strength values must be monotonically increasing.
They are also bounded from above by $1$ due to the nature of the differential equations, and so they must eventually converge.
For cycles with only attacks, things are already less straightforward because the attraction force of the initial weight
may become stronger than the attacking force as the attackers become weaker. This may let the strength
oscillate between the initial weight and the first lower peak reached. However, since arguments' strength can never exceed 
the initial weight if there is no support, the amplitude of the oscillations must eventually go to $0$
and the strength values will converge.

If cycles contain both support and attack relations, the strength values may oscillate more radically. 
In order to illustrate this, Figure \ref{fig_oscillation_example_bag} shows a BAG from a family that we call Cycle(k).
Each member contains one argument $A$ with initial weight $1$ that supports $k$ arguments $B_i$ with weight $0$.
Each $B_i$ in turn supports the same $k$ arguments $C_i$ that have initial weight $0$ as well. Finally, each $C_i$ attacks $A$. Figure \ref{fig_oscillation_example_bag}
shows Cycle(3) and figure \ref{fig_oscillation_example_graph} shows the long-term behaviour of the quadratic energy model for Cycle(3) at the top and for Cycle(10) at the bottom.
\begin{figure}[tb]
		  \scalebox{.68}{
				\xymatrix{
			%   Line1
 *++[o][F-]{B_1: 0} \ar@{-->}[r] \ar@{-->}@/^2pc/[rrr]  \ar@{-->}@/^2pc/[rrrrr] & *++[o][F-]{C_1: 0} \ar@{->}[ddr] & 
 *++[o][F-]{B_2: 0} \ar@{-->}[r] \ar@{-->}[l] \ar@{-->}@/^2pc/[rrr] & *++[o][F-]{C_2: 0} \ar@{->}[ddl] & 
 *++[o][F-]{B_3: 0} \ar@{-->}[r] \ar@{-->}[l] \ar@{-->}@/^2pc/[lll] & *++[o][F-]{C_3: 0} \ar@{->}[ddlll] \\
\mbox{}\\
  &&*++[o][F-]{A: 1} \ar@{-->}[uull] \ar@{-->}[uu] \ar@{-->}[uurr]\\
			}	
		}
		\caption{Cycle(3) graph. \label{fig_oscillation_example_bag}}
\end{figure}
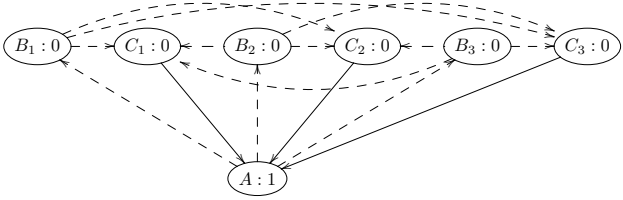
\begin{figure}[tb]
	\centering
		\includegraphics[width=0.46\textwidth]{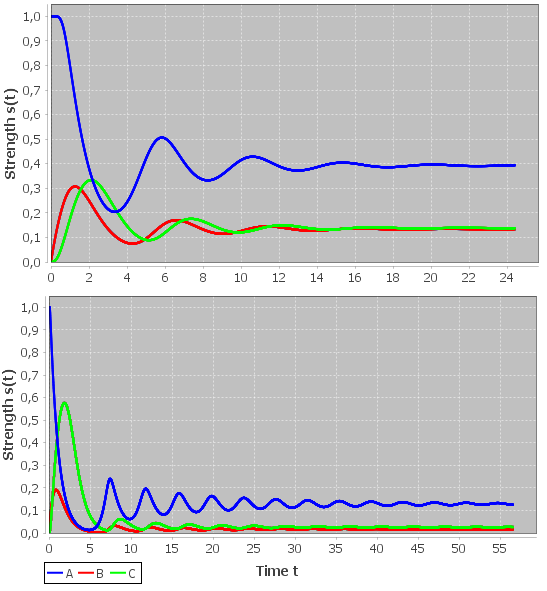}
			\caption{Long-term behaviour of quadratic energy model for Cycle(3) (top) and Cycle(10) (bottom).
			\label{fig_oscillation_example_graph}}
\end{figure}
As we may expect, the oscillations for Cycle(k) take more time as we increase $k$. However, the amplitude decreases and the strength values eventually converge.
It is currently unclear if there exist BAGs where the strength values oscillate for all time.  
However, experiments in \cite{potyka2018Kr}
with 3,000 randomly generated BAGs with thousands of nodes and ten thousands of edges
demonstrate that the quadratic energy model converges for many cyclic BAGs.

How can we deal with potential divergence in practice?
Since equilibrium states can usually be computed in seconds,
it is pragmatic to set a time limit for the quadratic energy model.
Say, if the model did not converge after $30$ seconds, the algorithm stops.
Arguments whose strength value has not converged yet can then be detected automatically 
because we must have $| \diff{s_j} | > \epsilon$ at such an argument.
The evolution of the strength value can then be plotted like in 
Figure \ref{fig_stock_example_graph} in order to see whether the strength value
diverges, oscillates with decreasing amplitude or just converges slowly for other reasons. 
In particular, even if the strength diverges,
it may oscillate between meaningful bounds. For example, if the strength oscillates
between $0.8$ and $0.9$, we could still infer that the argument is rather strong.
Of course, this analysis can also be performed automatically by just
storing lower and upper bounds and monitoring the derivatives (oscillations occur
when the derivative changes signs repeatedly).
However, since no non-convergent example has been found so far, 
we do not discuss these issues further here.  

Currently, my feeling is that the quadratic energy model always converges.
This assumption is based on the idea that the strength of every argument will reach a peak at some point in time.
After all arguments have reached their peak, I assume that the amplitude of oscillations will necessarily decrease 
similar to the observation in Figure \ref{fig_oscillation_example_graph}.
Intuitively, the overall energy in the system increases up to one point, but will necessarily decrease after having
reached its peak.
%More formally, I believe that the following conjecture is true. Of course, the reader is invited to prove this conjecture
%true or false.
%\todo[inline]{conjecture is false, e.g., supported argument will increase monotonically and never reach its peak - it is bounded, though}
%\begin{conjecture}
%\label{convergence_conjecture}
%\begin{enumerate}
	%\item For all arguments $i$ and all $\epsilon >0$, there is a time point $T_i$ such that $s_i(T_i) \geq s_i(t) - \epsilon$ for all $t \geq T_i$.
	%\item Starting from time $T = \max\{T_1, \dots, T_n\}$, the strength of every argument will be between bounds
	%$$0 \leq l_i(T) \leq l_i(t) \leq u_i(t) \leq u_i(T) \leq 1$$
	%for all $t \geq T$ and $| l_i(t) - u_i(t) | \rightarrow 0$ as $t\rightarrow \infty$.
%\end{enumerate}
%\end{conjecture}

On the bright side, while convergence for cyclic BAGs is a difficult question,
the quadratic energy model is guaranteed to converge for arbitrary acyclic BAGs. 
The equilibrium can be computed by numerical methods as discussed before,
but can also be computed by a discrete iteration scheme in linear time. 
The key observation is that arguments' strength depends only on the initial
weight and the strength of their parents (attackers and supporters).
By evaluating the arguments according to a topological
ordering, we can make sure that the final strength of all parents is known
in advance and we can compute the final strength values for every argument
in a single pass through the graph.
This is basically the same mechanism that is used to compute the strength values
for discrete models for weighted bipolar argumentation like in \cite{baroni2015automatic,rago2016discontinuity,amgoud2017evaluation}.  
We only present
the main result here and refer to \cite{potyka2018Kr} for more details and the proof. 
\begin{proposition}[Equilibria in Acyclic BAGs \cite{potyka2018Kr}]
\label{prop_acyclic_iteration_scheme}
Let $\bag$ be an acyclic BAG. Then $\energysolution$ converges and the equilibrium $s^* = \lim_{t \rightarrow \infty} \energysolution(t)$ can be computed in linear time by the following procedure:
\begin{enumerate}
  \item Compute a topological ordering of the arguments.
	\item Pick the next argument $i$ in the order and set 
	$$s_i = \weight(i) 
+  (1 - \weight(i)) \cdot h(E_i)  
- \weight(i) \cdot h(-E_i),$$
where $E_i$ is the energy at $i$.
	\item Repeat step 2 until all strength values have been computed.
\end{enumerate}
\end{proposition}
\begin{example}
We illustrate Proposition \ref{prop_acyclic_iteration_scheme} with an 
e-democracy problem from \cite{rago2016discontinuity}.
The question is how to spend a portion of a council's budget.
The arguments are divided into decision arguments (prefix A), pro arguments (prefix P)
and contra arguments.. 
\begin{description}
 \item[A1:] Build a new cycle path.
 \item[A2:] Repair current infrastructure.
 \item[P1:] Cyclists complain of dangerous roads.
 \item[P2:] A path would enhance the council’s green image.
 \item[P3:] Potholes have caused several accidents recently.
 \item[C1:] Significant disruptions to traffic would occur.
 \item[C2:] Environmentalists are a fraction of the population.
 \item[C3:] Recent policies already enhance this green image.
\item[C4:] Donors do not see the environment as a priority.
\end{description}
Figure \ref{fig_edemocracy_example_bag} shows the initial weights
and final strength values under the quadratic energy model. 
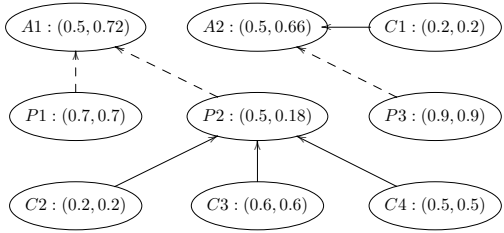
\begin{figure}[tb]
		  \scalebox{.65}{
				\xymatrix{
			%   Line1
 *+++[o][F-]{A1: (0.5,0.72)}  & *+++[o][F-]{A2: (0.5,0.66)} & 
*+++[o][F-]{C1: (0.2,0.2)} \ar@{->}[l]\\
*+++[o][F-]{P1: (0.7,0.7)} \ar@{-->}[u] &  *+++[o][F-]{P2: (0.5,0.18)} \ar@{-->}[ul] &
*+++[o][F-]{P3: (0.9,0.9)} \ar@{-->}[ul]  \\
*+++[o][F-]{C2: (0.2,0.2)} \ar@{->}[ur] & *+++[o][F-]{C3: (0.6,0.6)} \ar@{->}[u] &
*+++[o][F-]{C4: (0.5,0.5)} \ar@{->}[ul]     
			}	
		}
		\caption{BAG for e-democracy examples. Nodes show (initial weight, final strength) under
		quadratic energy model. \label{fig_edemocracy_example_bag}}
\end{figure}
Since the authors in \cite{rago2016discontinuity} considered only subgraphs
of this BAG, I defined additional initial weights for C2, C3, C4.
One topological ordering of the arguments is P1, P3, C1, C2, C3, C4, P2, A1, A2.
Since only P2, A1, A2 have parents, the energy at all other arguments is $0$ for all time
and their final strength is
just the initial weight. For P2, the energy is then $-0.2-0.6-0.5=-1.3$
and the final strength is 
$0.5 - 0.5 \cdot h(-(-1.3)) \approx 0.186$.
The energy at A1 is approximately $0.7+0.186 \approx 0.886$
and the final strength is 
$0.5 + 0.5 \cdot h(0.886) \approx 0.719$.
Finally, the energy at A2 is $0.9-0.2=0.7$ and 
the final strength is 
$0.5 - 0.5 \cdot h(0.7) \approx 0.664$.
The continuous evolution of the quadratic energy model is shown in Figure \ref{fig_edemocracy_example_graph}. Note that it does indeed converge to the values that we computed.
\begin{figure}[tb]
	\centering
		\includegraphics[width=0.46\textwidth]{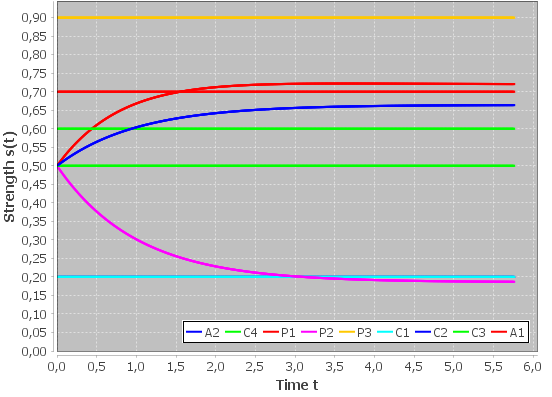}
			\caption{Long-term behaviour of $\energysolution$ for BAG in Figure \ref{fig_edemocracy_example_bag}.
			\label{fig_edemocracy_example_graph}}
\end{figure}
\end{example}
Even though we currently cannot give convergence guarantees for cyclic BAGs,
the quadratic energy model uniquely defines a strength value for every time point $t$.
In particular, we can analyze the runtime for evaluating the quadratic energy model
from time $0$ to time $T$.
Using step size $\delta$, this can be done in time $O(\frac{(\mid\arguments\mid + \mid\attacks\mid + \mid\supports\mid) \cdot T}{\delta})$.
The cost is basically composed of the factor 
$O((\mid\arguments\mid + \mid\attacks\mid + \mid\supports\mid))$
for the cost of evaluating the differential equations
and the factor
$O(\frac{T}{\delta})$
for the number of evaluations.
While Euler's method needs only a single computation of the differential equations
at each time step,
more sophisticated methods like RK4 evaluate the differential equations at several
points in order to improve the approximation.
While this increases the runtime for fixed $\delta$, these methods 
can usually work with significantly larger step sizes than Euler's method
and are therefore more efficient.
In our implementation, we let $T$ grow until $\| \diff{s} \|_\infty < 10^{-4}.$
It is reasonable to assume that the point of convergence $T^*$ depends on the
size and complexity of cycles. 
Experiments in \cite{potyka2018Kr} indicate that the overall runtime is bounded from above quadratically
by the size of the BAG. 

\section{From Discrete to Continuous Models}
\label{sec_discrete_continuous}

Proposition \ref{prop_acyclic_iteration_scheme} basically tells us
that if the BAG is acyclic,
 we can transform the continuous quadratic energy model to a
discrete model similar to the ones considered in
\cite{baroni2015automatic,rago2016discontinuity,amgoud2017evaluation}.
This is interesting from a computational perspective because it
gives us a linear runtime guarantee for acyclic BAGs.

On the other hand, continuous models are computationally interesting
because they can improve stability in cyclic BAGs as we explain at the end of this section.
Therefore, it is natural to ask under what conditions we can transform
discrete models like in 
\cite{baroni2015automatic,rago2016discontinuity,amgoud2017evaluation}
to well-defined continuous models. 
A simple sufficient criterion along with some guarantees is given in the following result from \cite{potyka2018Kr}.
\begin{proposition}[Continuizing Iterative Schemes]
\label{prop_continuizing_discrete_model}
Consider an iterative scheme $\iterscheme$ that defines the strength values for acyclic BAGs by letting
$$s_i = f_\iterscheme(\weight(i), \{s_j \mid j \in \attacker_i\}, \{s_j \mid j \in \supporter_i\}),$$
where $f_\iterscheme$ is a function that depends on the initial weight and the strength of attackers and supporters
and the strength values $s_i$ are computed in topological order.
\begin{enumerate}
	\item If $f_\iterscheme$ is continuously differentiable with respect to all involved strength values, then for all
	BAGs $\bag$, the system 
\begin{align*}
\diff{s_i} = f_\iterscheme(\weight(i), \{s_j \mid j \in \attacker_i\}, \{s_j \mid j \in \supporter_i\}) - s_i
\end{align*}
with initial conditions 
$s_i(0) = \weight(i)$ for $i = 1,\dots, n$ has a unique solution $\energysolution_\iterscheme: \mathbb{R}^+_0 \rightarrow \mathbb{R}^n$.
\item If $\energysolution_\iterscheme$ reaches an equilibrium state $s^* = \lim_{t \rightarrow \infty} \energysolution_\iterscheme(t)$, then
$s^*_i = f_\iterscheme(\weight(i), \{s^*_j \mid j \in \attacker_i\}, \{s^*_j \mid j \in \supporter_i\})$.
\item $\energysolution_\iterscheme$ reaches an equilibrium state whenever $\bag$ is acyclic.
\end{enumerate}
\end{proposition}
Item 1 explains how to transform the definition of a discrete iteration scheme
to a system of differential equations and
gives a sufficient condition under which the system
has a unique solution. We can then use this solution similar to the quadratic energy model as
we will illustrate soon.

Item 2 guarantees that if the model reaches an equilibrium state,
this state is a fixed point of the discrete update function $f_\iterscheme$.
This implies, in particular, that if the BAG is acyclic, then the continuized
model agrees with the discrete model.

Item 3 states that the continuized model is again guaranteed to convergence for acyclic graphs.

\subsection{Continuizing Discrete Models}

Let us now illustrate Proposition \ref{prop_continuizing_discrete_model} by
means of the Euler-based restricted semantics that was introduced in 
\cite{amgoud2017evaluation}.
As explained before, the Euler-based restricted semantics used the energy
$E_i = \sum_{i \in \supporter} s_i - \sum_{i \in \attacker} s_i$ before.
Given an acyclic BAG, the weights for every argument are then set in topological 
order by letting
\begin{equation}
s_i = 1 - \frac{1 - \weight(i)^2}{1 + \weight(i) \cdot \exp(E_i)}.
\label{eq:amgoud2017}
\end{equation}
That is, $f_\iterscheme(\weight(i), \{s_j \mid j \in \attacker_i\}, \{s_j \mid j \in \supporter_i\}) = 1 - \frac{1 - \weight(i)^2}{1 + \weight(i) \cdot \exp(E_i)}$.
In order to apply item 1, we have to check that $f_\iterscheme$ is continuously differentiable with respect to all $s_i$. Note that $E_i$ is a linear function of the
strength values and therefore continuously differentiable. The exponential function 
$\exp$ is continuously differentiable as well.
Therefore, $f_\iterscheme$ is defined by combining constant and continuously
differentiable functions and is therefore itself continuously differentiable
(notice, in particular, that the denominator in the fraction in $f_\iterscheme$
is always greater than $1$ because $\exp$ is a positive function).
Hence, we can apply Proposition \ref{prop_continuizing_discrete_model}.

Item 1 tells us that we obtain the differential equations for $s_i$ by
subtracting $s_i$ from $f_\iterscheme$.
Hence, the system that defines the continuous Euler-based semantics is
\begin{align}
\label{eq:euler_differential_equations}
\diff{s_i} = 1 - \frac{1 - \weight(i)^2}{1 + \weight(i) \cdot \exp(E_i)} - s_i, \qquad i \in \arguments.
\end{align}
We can now approximate the solution with Euler's method as described in Algorithm
\ref{fig:EulersMethodAlgorithm} or with faster methods like RK4.

Proposition \ref{prop_continuizing_discrete_model} is not always applicable. 
For example, the update formula for the DF-QuAD algorithm from \cite{rago2016discontinuity}
is not continuously differentiable.
The formula is also based on some auxiliary functions.
We slightly change the notation in order to make the presentation more homogeneous.
We define the \emph{geometric energy} at argument $j$
as 
$$\textit{GE}_j = \prod_{i \in \attacker_j} (1-s_i) - \prod_{i \in \supporter_j} (1-s_i),$$
where we use the convention that the empty product equals $1$. 
Given an acyclic BAG, the DF-QuAD algorithm sets the weights for every argument in topological 
order by letting
\begin{equation*}
s_i = 
\weight(i) 
+ \weight(i) \cdot \min\{\textit{GE}_i, 0\}
+ (1 - \weight(i)) \cdot \max\{\textit{GE}_i, 0\}.
\end{equation*}
We have $f_\iterscheme(\weight(i), \{s_j \mid j \in \attacker_i\}, \{s_j \mid j \in \supporter_i\}) =\weight(i) 
+ \weight(i) \cdot \min\{\textit{GE}_i, 0\}
+ (1 - \weight(i)) \cdot \max\{\textit{GE}_i, 0\}$.
The derivative of $f_\iterscheme$ is discontinuous at 0-energy states. 
Hence, Proposition \ref{prop_continuizing_discrete_model} is not applicable.
However, the conditions in  Proposition \ref{prop_continuizing_discrete_model}
are sufficient and not necessary. 
Indeed, one can show in another way that the system for the continuous DF-Quad algorithm 
has a unique solution \cite{potyka2018convergence}.

If Proposition \ref{prop_continuizing_discrete_model} is not applicable, we may also modify the update formula in order to guarantee continuous differentiability.
For the DF-Quad algorithm, we could square the strength values in the geometric energy.
When replacing the geometric energy with the  \emph{squared geometric energy} 
$$\textit{SGE}_j = \prod_{i \in \attacker_j} (1-s_i^2) - \prod_{i \in \supporter_j} (1-s_i^2),$$
Proposition \ref{prop_continuizing_discrete_model} is applicable.
When using the squared geometric energy,
an argument with strength $1$ will have the same influence as before, but as the strength gets
closer to $0$ the influence will get gradually weaker. Implementations of all continuizations can be found in 
\emph{Attractor} that we describe in the final section of this article.

\subsection{Continuization and Convergence}

To get an intuition for why continuizing a discrete model may improve the
convergence behaviour in cyclic graphs, it is instructive to look at Euler's method again.
Suppose we apply Euler's method with (rather large) step size $\delta = 1$
to the system given in Proposition \ref{prop_continuizing_discrete_model}.
Then we update each strength value $s_i$ with $s_i \leftarrow s_i + 1 \cdot \diff{s_i}$ in every iteration. Hence, the update is just
\begin{align*}
&s_i + 1 \cdot \big(f_\iterscheme(\weight(i), \{s_j \mid j \in \attacker_i\}, \{s_j \mid j \in \supporter_i\}) - s_i\big) \\
= &f_\iterscheme(\weight(i), \{s_j \mid j \in \attacker_i\}, \{s_j \mid j \in \supporter_i\}).
\end{align*}
That is, we just update all strength values simultaneously with respect to the iterative update formula $f_\iterscheme$.
Hence, applying the discrete iteration scheme can be seen as a very coarse approximation of 
a continuous system. In the presence of cycles, these coarse steps may lead to divergence 
even when the continuous model $\energysolution_\iterscheme$ converges. 
Intuitively, this is because a large step size like
$\delta = 1$ can let us jump from the graph of
the true solution $\energysolution_\iterscheme$ to the graph of another solution for different initial conditions. 
 By choosing a smaller step size, we can avoid these jumps and make the procedure more stable. This is basically what we are doing when continuizing $\iterscheme$.

\section{The Java Library Attractor}
\label{sec_attractor}

Finally, we will discuss how the previous ideas can be put into practice using the Java library \emph{Attractor}.
A download link for the current version is given in the footnote\footnote{\scriptsize\url{https://www.researchgate.net/publication/326677792_Attractor_v01}}.
The latest code is available at sourceforge\footnote{\scriptsize\url{https://sourceforge.net/projects/attractorproject/}}.
\emph{Attractor} is work in progress and currently provides only a programming interface. However, in the future, a graphical user interface will be added.
\emph{Attractor} can be used to
\begin{enumerate}
	\item compute solutions with existing models,
	\item use base classes to implement new models and
	\item evaluate continuous models on benchmarks and randomly generated BAGs.
\end{enumerate}
We will discuss each use case in turn. 

\subsection{Computing Solutions}

\begin{figure}[tb]
	\centering
		\includegraphics[width=0.46\textwidth]{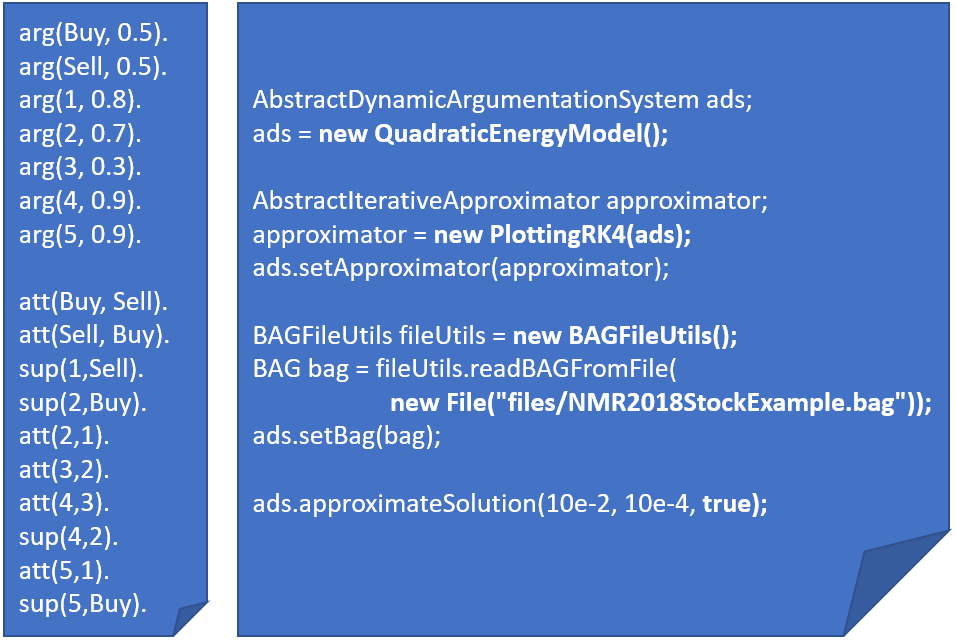}
			\caption{Reading a BAG from a file in \emph{Attractor}.\label{fig_Attractor_modelingExample}}
			
\end{figure}

BAGs can be created either programmatically or, more conveniently, by using a file reader. 
The programming approach is useful when considering families of BAGs like Cycle(k) (c.f. Figure \ref{fig_oscillation_example_bag} and \ref{fig_oscillation_example_graph}). 
A code example can be found in 
\emph{/Attractor/src/examples/NMR2018CycleK.java}. In this tutorial, we will focus on the file approach.
Figure \ref{fig_Attractor_modelingExample} shows an example file on the left and the code to compute the final strength values
with RK4 and to create a plot similar to Figure \ref{fig_stock_example_graph} on the right.
Files consist of definitions of arguments, attacks and supports. 
Argument definitions start with the keyword \emph{arg} and are followed by a name and an optional weight.
If no weight is provided in the definition, it is initialized with $0.5$ by default.
Attack and support definitions start with the keywords \emph{att} and \emph{sup} and are followed by the source
and the target of the edge as usual.
The file format is inspired by the format used in ConArg\footnote{\scriptsize\url{http://www.dmi.unipg.it/conarg/}} \cite{bistarelli2016conarg}, but adds
optional weights and support relations. The file reader in \emph{Attractor} can also read the current files from the ConArg benchmarks.

In order to compute a solution, we first have to initialize a model that is a subclass of the abstract class \emph{AbstractDynamicArgumentationSystem}
that we will explain in the next section.
The \emph{AbstractDynamicArgumentationSystem} reference \emph{ads} can also be initialized
with implementations of the continuized Euler-based semantics and DF-Quad algorithm
that we described in Section \ref{sec_discrete_continuous}.
By default, RK4 is used to compute solutions. Different algorithms can be selected by using the method \emph{setApproximator}.
In the example, we select \emph{PlottingRK4}, which still uses RK4, but simultaneously creates a plot for the evolution of the strength values
using JFreeChart\footnote{\scriptsize\url{http://www.jfree.org/jfreechart/}}.
The utility class \emph{BAGFileUtils} is used to read the file and the BAG object is passed to the model.
Afterwards, the call of the method \emph{approximateSolution} starts the approximation and will plot graphs like in Figure \ref{fig_stock_example_graph}.
The first two parameters determine the step size $\delta$ and the termination accuracy $\epsilon$. The third parameter is optional and can be used
to print the final strength values to the console. However, arguments and their strength values can also be accessed programmatically from the BAG object.

More file and programming examples can be found in \emph{Attractor/files} and \emph{Attractor/examples}. 
In particular, the code example \emph{NMR2018StockExampleComparison.java} shows how to compute 
and plot solutions for all models in \emph{Attractor} in order to compare the different semantics.

\subsection{Implementing new Models and Algorithms}

\begin{figure}[tb]
	\centering
		\includegraphics[width=0.46\textwidth]{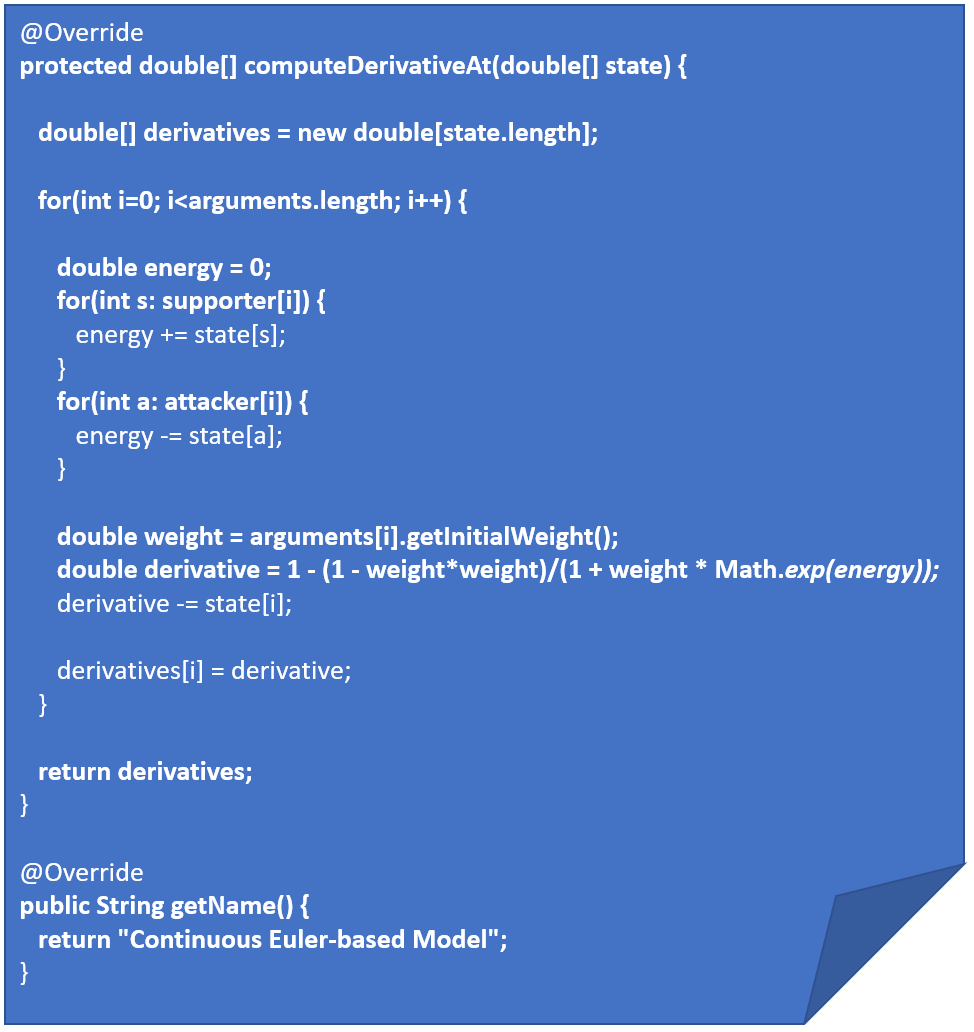}
			\caption{Implementation of Continuous Euler-based Model.\label{fig_Attractor_implementationExample}}
\end{figure}

New implementations of continuous models should be derived from the abstract class \emph{AbstractDynamicArgumentationSystem}
that can be found in the package \emph{edu.cs.ai.weightedArgumentation.dynamicalSystems}. The package already contains
implementations of the quadratic energy model and the continuized models that we discussed before.
\emph{AbstractDynamicArgumentationSystem} already provides most of the functionality, 
the programmer just has to implement the abstract methods \emph{computeDerivativeAt} that basically implements the differential equations 
and the method \emph{getName} that just returns the name of the model (this is used, for example, when creating plots).
Figure \ref{fig_Attractor_implementationExample} shows the implementation of the continuous Euler-based model.
The code in Figure \ref{fig_Attractor_implementationExample} is a straightforward translation of the derivatives given in Equation
\ref{eq:euler_differential_equations} into Java code.
As the code demonstrates, preinitialized arrays can be used to access arguments and their supporters and attackers efficiently.
All implementations of \emph{AbstractDynamicArgumentationSystem} can be used exactly as demonstrated in Figure \ref{fig_Attractor_modelingExample}.
In particular, different approximators can be selected. Currently, there are implementations of RK4, Euler's method
and a ploting variant of RK4. New algorithms should be derived from the abstract class
\emph{AbstractIterativeApproximator} and can then be selected analogously.

\subsection{Evaluating Models and Algorithms}

\begin{figure}[tb]
	\centering
		\includegraphics[width=0.46\textwidth]{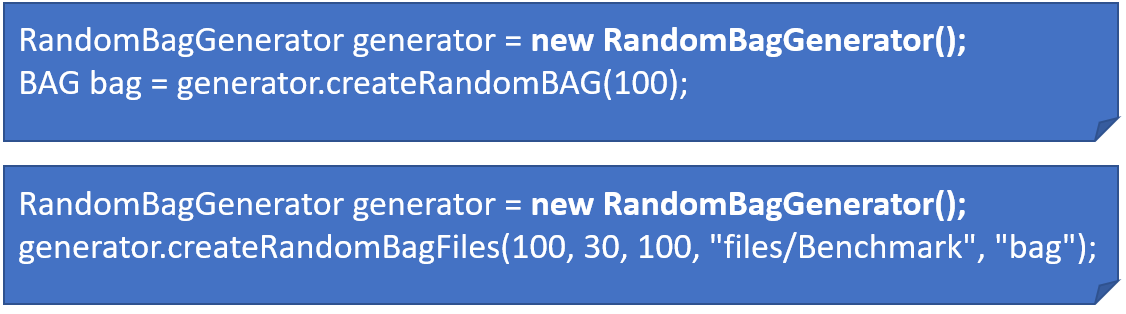}
			\caption{Generating a single random BAG (top) or a batch of random BAGs (bottom) in \emph{Attractor}.\label{fig_Attractor_randomgenerator}}
\end{figure}

Utility classes for evaluating models and algorithms can be found in the package \emph{edu.cs.ai.weightedArgumentation.util}.
The class \emph{RandomBagGenerator} contains the random generator used for creating the BAGs for the benchmark from \cite{potyka2018Kr}.
The original benchmark can be downloaded from the link given in the footnote\footnote{\scriptsize\url{https://www.researchgate.net/publication/326557254_Weighted_Bipolar_Argumentation_Benchmark_KR2018}}. Figure \ref{fig_Attractor_randomgenerator} shows how to create a single BAG of size 100 at the top and how to create a batch of BAGs of different sizes at the bottom.
The arguments for the method \emph{createRandomBagFiles} allow configuring the basic size (100), the number of increments of the size (30) and the number of trials for each size (100).
In the example in Figure \ref{fig_Attractor_randomgenerator}, 100 graphs of size $100, 200, 300, \dots, 3000$ each will be created and stored in the local directory 'files/Benchmark',
each file starting with the prefix 'bag'.

\begin{figure}[tb]
	\centering
		\includegraphics[width=0.46\textwidth]{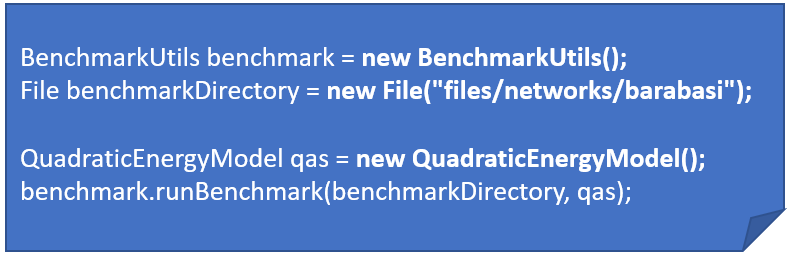}
			\caption{Running benchmarks in \emph{Attractor}.\label{fig_Attractor_benchmarks}}
\end{figure}

\begin{figure}[tb]
	\centering
		\includegraphics[width=0.46\textwidth]{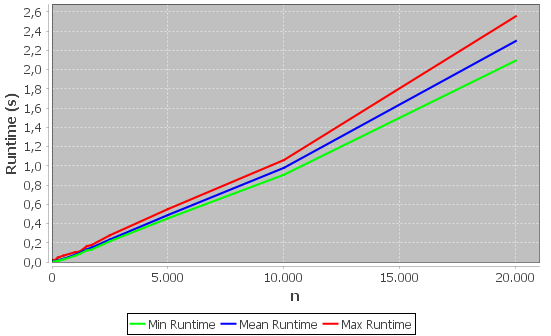}
			\caption{Runtime results for quadratic energy model model on ConArg Barabasi benchmark.\label{fig_conarg_barabasi}}
\end{figure}

The class \emph{BenchmarkUtils} can be used to run benchmarks and to plot statistics for the evaluation similar to the evaluation in \cite{potyka2018Kr}. 
Figure \ref{fig_Attractor_benchmarks} shows how to run the benchmark files in a directory. 
The method \emph{runBenchmark} assumes that the given directory contains subdirectories. Each of these subdirectories contains BAGs of a fixed size and the name of 
the directory is supposed to be the size. The passed model will then be evaluated on all benchmark files and the method stores minimum, mean and maximum runtime
for all sizes. Runtime results for individual files are printed to the console. The statistics are plotted simultaneously as shown in Figure \ref{fig_conarg_barabasi}.
In this case, we evaluated the quadratic energy model on the Barabasi files from the ConArg benchmark. Other implementations of the base class \emph{AbstractDynamicArgumentationSystem} 
can be evaluated analogously. 
Let us note that the ConArg benchmark does not contain weights and supports. By default, all weights will be set to $0.5$.
In order to evaluate new models on BAGs with supports, the benchmark from \cite{potyka2018Kr} can be used or new 
benchmarks can be generated using the class \emph{RandomBagGenerator}.

\section{Conclusions and Future Work}

Continuous dynamical systems are an alternative to discrete models that may give stronger convergence guarantees 
for cyclic BAGs in the future. Analyzing the general convergence behaviour is difficult and there are no general
convergence guarantees currently. However, experiments show that continuous models converge in many cyclic
BAGs and do so quickly. There are also interesting relationships between continuous and discrete models.
For acyclic BAGs, equilibrium states are guaranteed to exist and can be computed by a discrete iteration scheme.
This is computationally advantageous because it gives us a linear runtime guarantee.
While continuous models converge superlinearly, they converged subquadratically in all previous experiments
and the ability to plot the continuous evolution of strength values may be interesting to improve the
explainability of the final strength values even for acyclic graphs.
Existing discrete models can be transformed to continuous iteration schemes and continuous
differentiability of the update formula is a sufficient condition for some basic guarantees.

In order to simplify the use of continuous models, \emph{Attractor} provides basic implementations of the ideas
discussed here and in \cite{potyka2018Kr}. For applications, it allows computing solutions for weighted
argumentation problems. For further development, it allows deriving new models and algorithms from base classes
that already provide basic functionality. In particular, utility functions can be used to evaluate new models
and to compare them to existing models.

One main goal of future work is to advance the understanding of convergence conditions in cyclic BAGs.
This involves trying to prove convergence in general cyclic BAGs or finding a counterexample.
Furthermore, some empirical studies on the applicability in decision support and the analysis of Twitter
discussions similar to the work in \cite{baroni2015automatic,rago2016discontinuity,alsinet2017weighted} shall be conducted.
Developing a graphical user interface for \emph{Attractor} to simplify experiments
 will also be part of future work. 

%\bibliographystyle{plain}
%\bibliographystyle{abbrv}
% LNCS style
\bibliographystyle{aaai}  % Springer-Vorgabe, egibt alphabetische Sortierung

\bibliography{biblio}

\end{document}